\documentclass[9pt,shortpaper,twoside,web]{ieeecolor2}
\usepackage{generic}
\usepackage{cite}
\usepackage{amsmath,amssymb,amsfonts}
\usepackage{graphicx}
\usepackage{textcomp}
\def\BibTeX{{\rm B\kern-.05em{\sc i\kern-.025em b}\kern-.08em
    T\kern-.1667em\lower.7ex\hbox{E}\kern-.125emX}}
\usepackage[table,x11names,dvipsnames]{xcolor}
\usepackage{multirow} 
\usepackage{booktabs}
\usepackage{hyperref}
\usepackage{amssymb}
\usepackage{amsmath}
\usepackage{acronym}
\usepackage{subcaption}
\usepackage{algorithm}
\usepackage{algpseudocode}

\newcommand{\LineComment}[1]{%
    \Statex \textcolor{OliveGreen}{#1}%
}

\begin{document}

\title{
% Bridging Discrete Tokens and Continuous Quantitative Prediction in Medical Vision-Language Models by Preserving Numerical Continuity
Beyond Discrete Tokens: Preserving Numerical Continuity\\ in Medical Vision-Language Models
% HAD-VLM: Bridging Discrete Language Generation and Continuous Quantitative Prediction in Medical Vision-Language Models
% Preserving Numerical Continuity in Medical Vision-Language Models
% When Number Matters: Preserving Numerical Continuity in Medical Vision-Language Models
}

\author{Dianye Huang, Jiamei Guo, Yue Zhou, Mingde Li, \\Yi Xiong, Wei Wang, Nassir Navab, and Zhongliang Jiang 
% \IEEEmembership{Member, IEEE}
% \thanks{This paragraph of the first footnote will contain the date on 
% which you submitted your paper for review. It will also contain support 
% information, including sponsor and financial support acknowledgment. For 
% example, ``This work was supported in part by the U.S. Department of 
% Commerce under Grant BS123456.'' }
\thanks{Dianye Huang and Jiamei Guo contributed equally to this work. (Corresponding author: Zhongliang Jiang.)}
\thanks{Dianye Huang and Zhongliang Jiang are with the Medical Intelligence and Robotic Cognition (MIRoC) Lab, Department of Mechanical Engineering, The University of Hong Kong, Hong Kong SAR, China (dianye.huang@hku.hk; zljiang@hku.hk).}
\thanks{Jiamei Guo, Yue Zhou and Nassir Navab are with the Chair for Computer-Aided and Medical Procedures (CAMP), Tehcnical University of Munich, Munich, Germany (jiamei.guo@tum.de;yue.zhou@tum.de; nassir.navab@tum.de).}
\thanks{Mingde Li is with the Department of Medical Ultrasonics, The First Affiliated Hospital of Sun Yat-Sen University, Guangzhou, China. (mingdeli@yeah.net)}
\thanks{Yi Xiong is with Shenzhen Luohu People's Hospital, Shenzhen, China. (13352995536@163.com)}
\thanks{Wei Wang is with the Department of Medical Ultrasonics, Guangdong Provincial People's Hospital, Guangzhou, China. (ultrasonography@live.com)}
}

% - dianye.huang@tum.de
% - jiamei.guo@tum.de
% - limd25@mail2.sysu.edu.cn
% - wangw73@mail.sysu.edu.cn
% - nassir.navab@tum.de
% - zljiang@hku.hk

% 【1】Mingde Li
% E-mail: mingdeli@yeah.net
% Department of Medical Ultrasonics, The First Affiliated Hospital of Sun Yat-Sen University, Guangzhou, China

% 【2】Wei Wang
% E-mail: ultrasonography@live.com
% Department of Medical Ultrasonics, Guangdong Provincial People's Hospital, Guangzhou, China

\maketitle

\begin{abstract}
\textit{Objective:}
To develop a unified vision-language model for prompt-conditioned quantitative ultrasound measurement that addresses the mismatch between continuous measurement targets and discrete token-based numerical prediction.
\textit{Methods:} This article presents HAD-VLM, a hybrid autoregressive-diffusion framework that combines autoregressive text generation with continuous measurement prediction within a shared backbone. Visual prompt tuning adapts the vision encoder to ultrasound images. A diffusion process predicts a five-component representation comprising caliper endpoint coordinates and measurement length. A post-processing strategy combining the proposed coordinate-length consistency mechanism with ensemble prediction is introduced to refine measurement estimates. We evaluate four echocardiographic measurement tasks on EchoNet-LVH, with additional assessments of medical question answering and X-ray vertebral measurement.
\textit{Results:}
HAD-VLM achieves mean absolute errors of 1.36, 2.77, 3.05, and 1.16 mm for IVSd, LVIDd, LVIDs, and LVPWd, respectively, reducing errors by 21.4-37.6\% relative to the Vicuna-based autoregressive baseline. It also outperforms fine-tuned Qwen3-VL and direct MLP regression across all four tasks. Medical question answering performance is preserved on the evaluated dataset. For X-ray vertebral measurement, HAD-VLM reduces mean absolute error from 11.4 to 6.7 pixels.
\textit{Conclusion:}
Integrating autoregressive language generation with diffusion-based continuous prediction improves quantitative medical image measurement while retaining prompt-conditioned task flexibility.
\textit{Significance:}
HAD-VLM provides a unified approach to combining language interaction with anatomically grounded numerical prediction for quantitative medical image analysis.

\end{abstract}

\begin{IEEEkeywords}
% Ultrasound Image; Vision-Language-Model; Ultrasound Measurement.
Vision-Language Models, Ultrasound Image, Ultrasound Measurement, Multi-task Learning.
% Enter key words or phrases in alphabetical 
% order, separated by commas. For a list of suggested keywords, send a blank 
% e-mail to keywords@ieee.org or visit \underline{$\small
% {http://www.ieee.org/organizations/pubs/ani\_prod/keywrd98.txt}$}
\end{IEEEkeywords}

% submission guideline: https://www.embs.org/tbme/prepare-a-manuscript/

% Size: All the text, figures, and tables should be in one file with a size of less than 10 MB for review purposes. 

% Cover Letter: All submissions must include a Cover Letter of less than 250 words, to be submitted in the text box when submitting your manuscript via the IEEE Author Portal.
% Abstract: It must be less than 250 words with the following subheads: objective, methods, results, conclusion, and significance.  a one–sentence statement of significance to biomedical research.
% Conclusion: Each TBME article must include a conclusion section that elaborates on the major findings and significance of the work, as well as its applications. This section should not exceed 300 words. 

% Biography: Beginning with the January 2018 issue, TBME will no longer include author biographies or author photographs. 

% Submission: Authors are strongly encouraged to suggest 4 potential reviewers, although they may not be requested to review the submission. 

% Review process: using a single-anonymous peer review process,

% Page Limit: Eight (8) or fewer pages are expected for a regular manuscript. While a regular manuscript can have a maximum length of 12 pages,

\section{Introduction}
\label{sec:introduction}
\par
Quantitative measurements are essential components of clinical image interpretation. Anatomical dimensions, lesion diameters, and ventricular volumes provide objective information for diagnosis, disease characterization, and longitudinal follow-up~\cite{li2024ultrasound,duffy2022high}. In ultrasound (US), obtaining these measurements often requires identifying anatomical landmarks and manually positioning calipers or tracing anatomical boundaries. These procedures increase examination workload and are subject to intra- and inter-observer variability~\cite{strauss2007interobserver,sabo2023real}. Deep learning has therefore been widely investigated to automate US measurement, with promising results in anatomical segmentation, landmark localization, and quantitative cardiac assessment~\cite{zhou2021review,ouyang2020video,duffy2022high}. Nevertheless, many existing systems operate on predefined anatomical targets and output schemas~\cite{zhang2015automatic,liang2025quasi,jafari2021u}. 

\begin{figure}[!t]
    \centering 
    \includegraphics[width=0.90\linewidth]{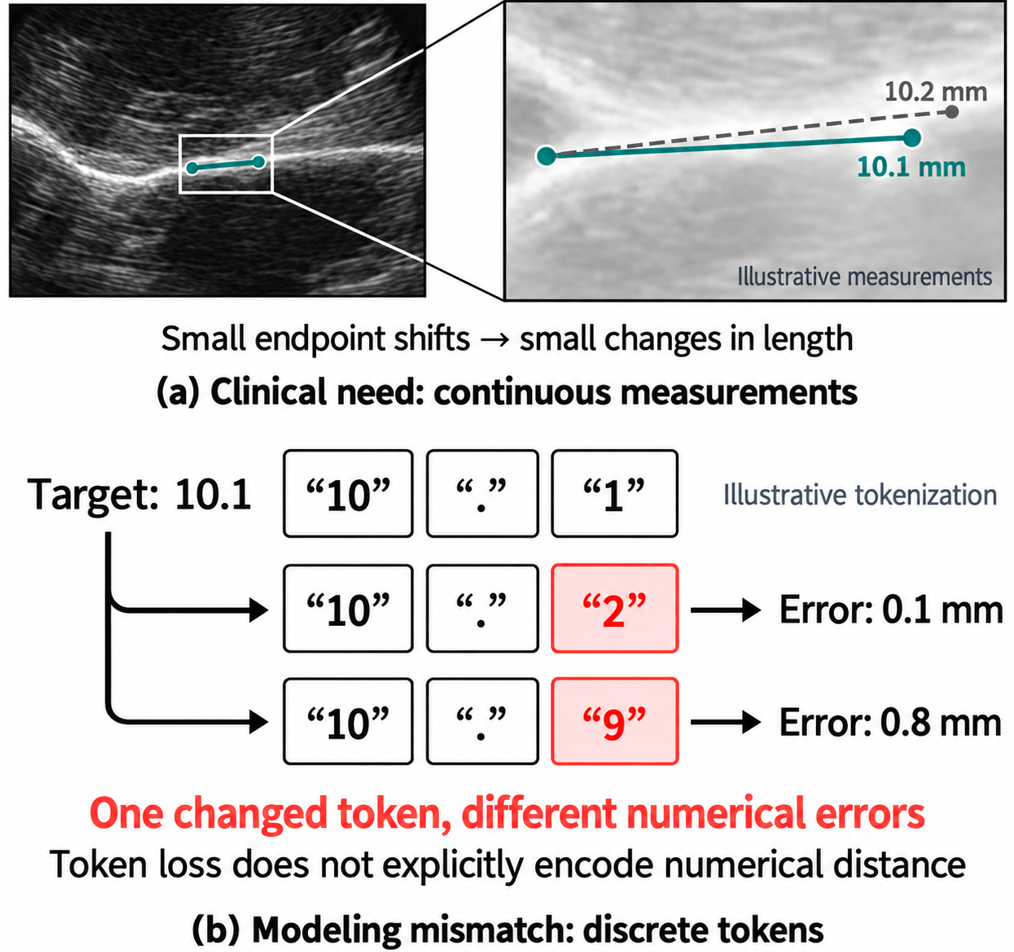}
    \caption{An illustration of the mismatch between continuous anatomical measurements and discrete token-based numerical prediction.}
    \label{fig:teaser}
\end{figure}

\par
Vision-language models (VLMs) offer a complementary approach by allowing image analysis tasks to be specified through natural-language instructions. Advances in biomedical visual question answering and radiology report generation demonstrate the potential of a shared vision-language backbone to support diverse image interpretation tasks~\cite{li2023llava,bannur2024maira,li2025medical}. For quantitative US assessment, such an interface could allow a clinician to specify the desired measurement through a text prompt while sharing visual and semantic representations across related tasks. Realizing this capability, however, requires accurate numerical prediction alongside language generation.

\begin{figure*}[!t]
    \centering 
    \includegraphics[width=0.95\linewidth]{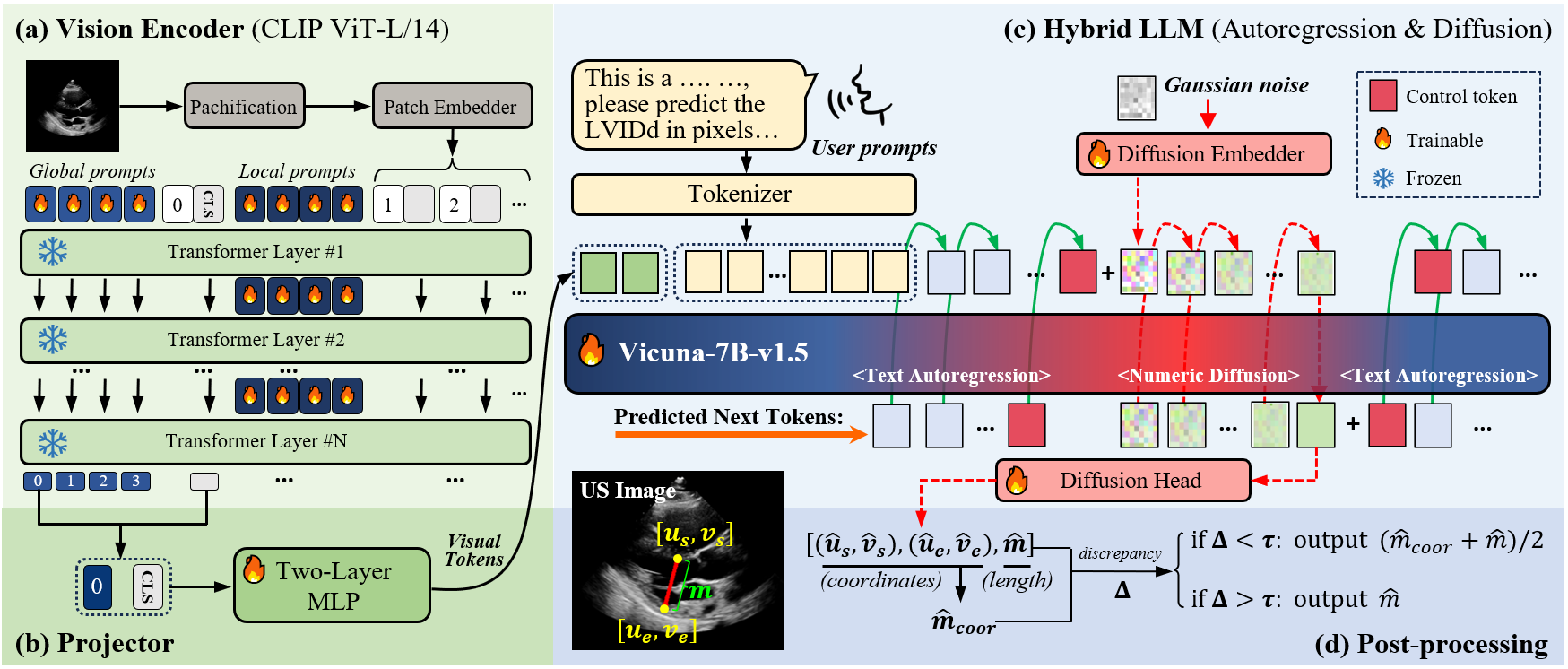}
    \caption{Overview of HAD-VLM for prompt-conditioned quantitative ultrasound measurement. The framework comprises (a) an prompt-adapted vision encoder, (b) a projector, (c) a hybrid LLM, and (d) a post-processing module. It integrates autoregressive text generation with diffusion-based continuous measurement prediction via a 5-tuple measurement representation and a coordinate-length consistency mechanism.}
    \label{fig:overall_framework}
\end{figure*}

\par
A central challenge lies in the mismatch between continuous measurement targets and discrete token-based output modeling, as illustrated in Fig.~\ref{fig:teaser}. Clinically relevant quantities, such as distances and diameters, are continuous variables, whereas autoregressive VLMs typically represent numerical values as token sequences and optimize their likelihood through categorical next-token prediction. Although language models can learn numerical relationships, this objective does not explicitly account for the magnitude of numerical errors~\cite{wallace2019nlp,xue2024reo}. For example, depending on the tokenizer, the value '10.1' may be decomposed into '10', '.', and '1', each generated sequentially. Such a representation does not explicitly preserve numerical proximity: similar values may have different tokenizations, and token-level prediction errors do not directly reflect distances between the corresponding values. For caliper-based measurements, small perturbations in endpoint coordinates should produce correspondingly small changes in the measured length; however, conventional token-based prediction does not explicitly enforce this geometric relationship. Errors in individual digits or decimal placement can also substantially alter the predicted value, potentially compromising measurement precision. Furthermore, token-by-token generation for inferring numerical values can introduce cumulative errors, hindering precise diagnostic measurements. 

\par
To address this challenge, we propose \textbf{HAD-VLM}, a hybrid autoregressive-diffusion vision-language model for prompt-conditioned quantitative US measurement. HAD-VLM combines autoregressive text generation with diffusion-based prediction of continuous measurements within a unified backbone. To connect numerical outputs with their spatial interpretation, we introduce a five-tuple measurement representation that couples caliper endpoint coordinates with the corresponding length, together with a coordinate-length consistency mechanism that encourages agreement between the predicted geometry and measurement value.

\par
Our objective is to support multiple quantitative measurement tasks through a common language-conditioned interface while retaining the numerical precision required for meaningful image assessment. The shared architecture enables task conditioning through text prompts and joint learning across measurement targets. 
% This study thereby focuses on quantitative measurement as a component of future language-enabled clinical workflows; comprehensive clinical report generation remains a direction for further development. 
The main contributions are summarized as follows:
\begin{itemize}
    \item We introduce HAD-VLM, a unified framework that combines autoregressive text generation with continuous diffusion-based prediction for prompt-conditioned US measurement.
    \item We develop a five-tuple measurement representation and a \textit{coordinate-length consistency} mechanism to couple anatomical localization with numerical estimation.
    \item We evaluate the framework across multiple US measurement tasks and analyze the effects of cross-task learning, continuous prediction, and the hybrid architecture on measurement accuracy and language performance.
\end{itemize}
To the best of our knowledge, this is the first work to introduce a hybrid autoregressive-diffusion formulation for quantitative US measurement, targeting precise numerical prediction while preserving the flexible language capabilities of VLMs within a single unified model. Specifically, we evaluate the framework in both ultrasound and X-ray imaging to examine its applicability across modalities. 
On ultrasound images, HAD-VALM performs four echocardiographic measurements specified through natural-language prompts, achieving measurement errors comparable in magnitude to previously reported inter-observer differences. For LVIDd and LVIDs, MAEs of 2.77 and 3.05~mm are numerically lower than the inter-observer differences of 2.97 and 3.60~mm reported in a separate study~\cite{Tromp2022AFV}. These findings support the potential of HAD-VLM for accurate, language-guided quantitative medical image assessment.

\section{Related Work}
\label{sec:related_work}
\subsection{Deep Learning for Quantitative Ultrasound Assessment}
\par
Deep learning approaches to quantitative US assessment commonly estimate anatomical landmarks, segment structures, or directly regress clinically relevant quantities~\cite{zhou2021review,SAHASHI2025}. In echocardiography, EchoNet-Dynamic combines ventricular segmentation with video-based estimation of cardiac function, enabling automated beat-to-beat assessment~\cite{ouyang2020video}. EchoNet-LVH further demonstrates automated measurement of ventricular dimensions and wall thickness for quantitative characterization of left ventricular hypertrophy~\cite{duffy2022high}. In lung ultrasound, Lucassen \emph{et al.} benchmark deep learning methods for B-line detection and introduce single-point localization of B-line origins to support interpretable quantification~\cite{lucassen2023deep}. These studies establish strong task-specific approaches to reducing manual measurement effort and improving reproducibility. Such systems may support several related measurements, but their requested outputs are generally specified by a predefined task configuration. Our work investigates a language-conditioned formulation in which the measurement target is specified through a prompt and processed by a shared model. This formulation aims to facilitate joint learning across measurement tasks and provide a common interface for anatomical localization, numerical estimation, and associated textual outputs. Specialized measurement models remain important references for assessing the accuracy of this more flexible formulation.

\subsection{Vision-Language Models for Medical Image Interpretation}
\par
Medical VLMs have expanded the scope of image analysis from fixed-label prediction to instruction-following and open-ended language generation. LLaVA-Med adapts a general-purpose vision-language assistant to biomedical images through domain-specific alignment and visual instruction tuning~\cite{li2023llava}. For echocardiographic video analysis, EchoAgent uses an LLM to orchestrate specialized vision tools for temporal localization, anatomical measurement, and guideline-based clinical interpretation~\cite{daghyani2026echoagent}. In radiology, MAIRA-2 generates chest X-ray reports with spatial grounding of findings, linking textual descriptions to supporting image regions~\cite{bannur2024maira}. These developments illustrate the value of integrating visual evidence and language within a shared model. Quantitative measurement introduces an additional requirement: numerical outputs must be accurate and consistent with the underlying image geometry. Success in visual question answering or report generation does not, by itself, establish this capability. Studies of numerical representations show that language models can encode magnitude information, but numerical precision depends on the representation and training objective~\cite{wallace2019nlp}. HAD-VLM addresses this requirement by explicitly modeling continuous measurement variables alongside autoregressive textual outputs and linking predicted lengths to their associated caliper coordinates.

\subsection{Continuous Prediction in Vision-Language Frameworks}
\par
Recent studies have explored continuous prediction within language-conditioned architectures. Xue~\emph{et al.} proposed REO-VLM that augments a VLM with a regression branch for Earth observation and incorporates language-derived information into numerical estimation~\cite{xue2024reo}. It demonstrates that generative and regression capabilities can coexist within a unified framework, providing a relevant precedent for quantitative vision-language modeling. Hybrid autoregressive-diffusion models offer another route to combining discrete and continuous outputs. Transfusion trains a shared Transformer using next-token prediction for text and diffusion for continuous image representations~\cite{zhou2025transfusion}. In robotics, HybridVLA integrates autoregressive and diffusion-based action prediction within a shared language-model backbone and combines their outputs for manipulation~\cite{liu2025hybridvla}. DiffusionVLA likewise combines autoregressive reasoning with diffusion-based action generation~\cite{wen2025diffusionvla}. Together, these studies demonstrate that language-related processing and continuous generation can be learned within coordinated architectures.

\par
HAD-VLM builds on this general modeling principle and adapts it to quantitative US measurement. Its task requires continuous predictions to agree with both the language-specified anatomical target and the geometry of the corresponding caliper placement. Accordingly, HAD-VLM couples endpoint coordinates and measurement length in a structured representation and introduces an explicit coordinate-length consistency mechanism. The distinction lies in this measurement-specific formulation and its integration with textual outputs. The benefit of diffusion over direct regression is assessed empirically through the corresponding ablation.

\section{Method}
\label{sec:method}
\par
To jointly generate ``discrete'' textual descriptions and ``continuous'' numerical values for US measurements, we propose HAD-VLM. In this model, each numerical value is predicted directly as a continuous variable via a diffusion-based process, with the VLM serving as the denoising backbone. As shown in Fig. \ref{fig:overall_framework}, the overall architecture comprises four main components: (a) a prompt-adapted vision encoder tailored to US data; (b) an MLP projector that aligns visual feature representations with the LLM token space; (c) an LLM augmented with a diffusion embedder and a diffusion head dedicated to numerical regression, and (d) a post-processing module that further refines the predicted measurements. Detailed descriptions of each component are provided below.

\subsection{Adapting Vision Encoder to US Data via Learnable Prompts} 
As shown in Fig.~\ref{fig:overall_framework} (a), we adopt CLIP ViT-L/14~\cite{radford2021learning} as the visual backbone and use Visual Prompt Tuning (VPT)~\cite{jia2022visual} to adapt its representations to US images while keeping the pretrained backbone frozen. Following DGL~\cite{yang2024dgl}, we introduce $L=4$ learnable local prompts at each Transformer layer and $G=4$ global prompts at the first layer. The global prompts propagate through subsequent layers, whereas the local prompts are layer-specific.

\par
Let $I$ denote the input image and $\mathcal{P}$ the collection of learnable prompts. The prompted encoder produces the final class-token representation $h_{\mathrm{cls}}^{(D)}$ and global-prompt representations $\{h_{g,j}^{(D)}\}_{j=1}^{G}$. We concatenate the class token and the first global-prompt token to obtain visual representation as follow
\begin{equation}
    f_v =
    \operatorname{Concat}\!\left(
        h_{\mathrm{cls}}^{(D)},h_{g,1}^{(D)}
    \right),
    \label{eq:visual_representation}
\end{equation}
where $\left\{h_{\mathrm{cls}}^{(D)},h_{g,j}^{(D)}\right\}\subset F_{\mathrm{CLIP}}(I;\mathcal{P})$, and $D$ is the number of Transformer layers.

\par
A two-layer MLP projects this representation into the LLM embedding space:
\begin{equation}
    \mathbf{v} = P_{\psi}(f_v)
      = W_2\,\operatorname{GELU}(W_1 f_v+b_1)+b_2,
    \label{eq:visual_projector}
\end{equation}
where $\psi=\{W_1,b_1,W_2,b_2\}$ denotes the projector parameters. The projector is initialized from LLaVA~\cite{liu2023visual}, subject to compatibility of its input feature dimensions.

\begin{figure}[!t]
    \centering 
    \includegraphics[width=0.95\linewidth]{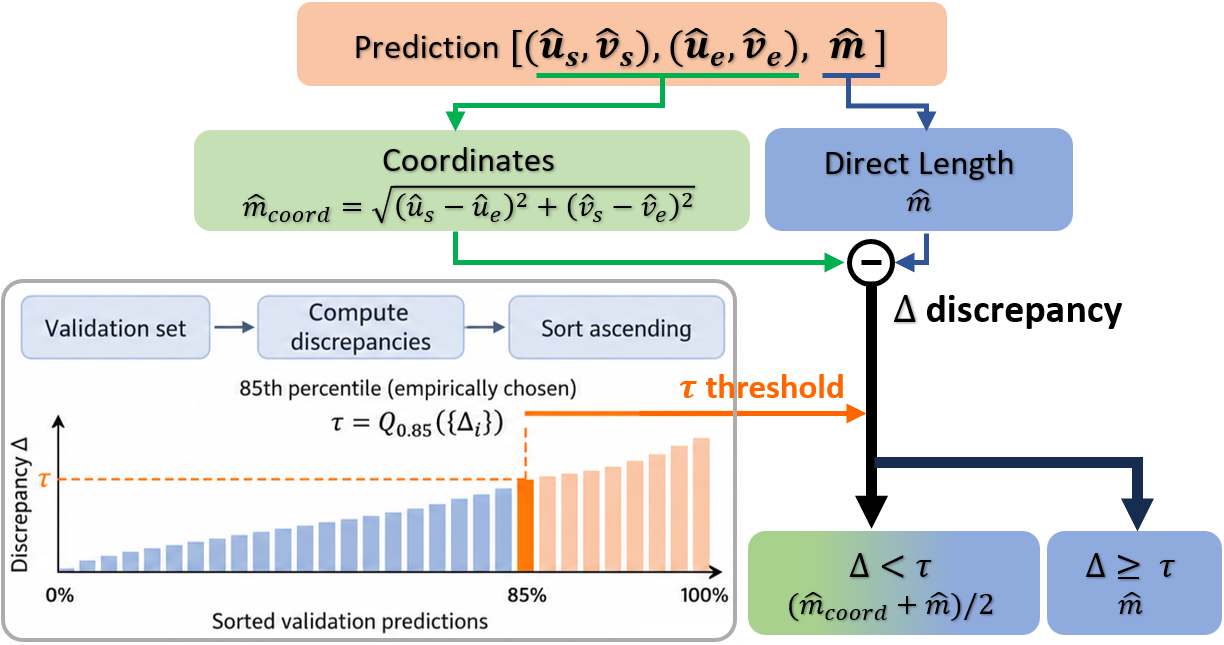}
    \caption{Illustration of the \textit{coordinate-length consistency} mechanism with a threshold selection based on validation-set discrepancies.}
    \label{fig:post_proc}
\end{figure}

\begin{algorithm}[!t]
\caption{HAD-VLM \textbf{Inference}}
\label{alg:hadvlm_inference}
\begin{algorithmic}[1]
\Require Image $I$, prompt $p$, trained model $\Theta$
\Require Ensemble size $R$, threshold $\tau$, decoding budget $N_{\max}$
\Require DDIM schedule $\{t_k\}_{k=0}^{K}$, $K=4$
% \Ensure Text $y$ and measurement records $\mathcal{M}$
\Ensure Text $y$ and measurement estimation $\hat{x}_0$

\State Encode $I$ to obtain $\mathbf{v}$ using
    Eqs.~\eqref{eq:visual_representation}--\eqref{eq:visual_projector}
\State Initialize empty sequences $y$

\For{$n=1,\ldots,N_{\max}$}
    \LineComment{\quad\texttt{//} Autoregressive text generation}
    \State Decode token $a$ from
        $P_{\Theta}(\cdot\mid\mathbf{v},p,y)$
    \If{$a=\texttt{<EOS>}$}
        \State \textbf{break}
    \EndIf
    \State $y\gets y\mathbin{\Vert}a$
        \Comment{Append the generated token}

    \LineComment{\quad\texttt{//} Diffusion-based quantitative measurement}
    \If{$y$ ends with the opening delimiter \texttt{[}}
        \State $c\gets
            \operatorname{P}_{\Theta}(\cdot\mid\mathbf{v},p,y)$
        \For{$r=1,\ldots,R$}
            \State $\hat{x}_0^{(r)}\gets
                \operatorname{DDIMSample}_{\theta}
                (c,\{t_k\}_{k=0}^{K})$
            \State $\tilde{m}^{(r)}\gets
                \operatorname{D}_{\tau}(\hat{x}_0^{(r)})$
        \EndFor

        \LineComment{\quad\quad\quad\texttt{//} Ensemble aggregation and sample selection}
        \State Compute $r^\star,~m^\star$ using Eq.~\eqref{eq:ensemble_median}
        % \State Append $(\hat{x}_0^{(r^\star)},m^\star)$
        %     to $\mathcal{M}$

        \LineComment{\quad\quad\quad\texttt{//} Insert measurement and resume text generation}
        \State $y\gets y\mathbin{\Vert}\operatorname{Tokenize}\bigl(\operatorname{String}(m^\star)\bigr) \mathbin{\Vert}\texttt{]}$
    \EndIf
\EndFor

\State \Return $\left(y,\hat{x}_0^{(r^\star)}\right)$
\end{algorithmic}
\end{algorithm}

\subsection{US Measurement with Hybrid LLM} 
\subsubsection{Model Architecture} 
\par
As illustrated in Fig.~\ref{fig:overall_framework} (c), HAD-VLM uses Vicuna-7B-v1.5~\cite{zheng2023judging} as its LLM backbone. The input comprises tokenized user prompts $p$ together with projected visual tokens $\mathbf{v}$ from the vision-language connector. The model produces two types of outputs: (1) textual descriptions generated autoregressively, and (2) numerical measurements predicted via a diffusion-based process. Each numerical measurement is represented as a continuous 5-tuple:
\begin{equation}
    x_0 =
    [u_s,v_s,u_e,v_e,m]^{\top}\in\mathbb{R}^{5},
    \label{eq:measurement_tuple}
\end{equation}
where $(u_s,v_s)$ and $(u_e,v_e)$ denote normalized start and end coordinates, respectively, and $m$ is the corresponding length in pixels. 

\par
This 5-tuple is regressed via a DDPM-based denoising process from $x_t$ to $x_0$ followed by a diffusion head~\cite{ho2020denoising}. A learnable embedder maps the noisy tuple and timestep into the LLM embedding space:
\begin{equation}
    z_t=\operatorname{E}_{\mathrm{diff}}(x_t,t).
    \label{eq:diffusion_embedding}
\end{equation}
This embedding is inserted at the numerical prediction slot. Let $c$ denote the preceding visual and textual context. The LLM processes the augmented sequence, and the diffusion head predicts the noise from the hidden representation at that slot:
\begin{equation}
    \hat{\epsilon}_{\theta}(x_t,t,c)
    =
    g_{\phi}\!\left(
        [\operatorname{P}_{\Theta}(c,z_t)]_{\mathrm{diff}}
    \right)
    \in\mathbb{R}^{5},
    \label{eq:noise_prediction}
\end{equation}
where $g_{\phi}(\cdot)$ is the diffusion head, $\operatorname{P}_{\Theta}(\cdot)$ is the LLM, and $\Theta$ denotes the model parameters. Under this noise-prediction parameterization, the clean tuple estimate at timestep $t$ is
\begin{equation}
    \hat{x}_{0\mid t}
    =
    \frac{x_t-\sqrt{1-\bar{\alpha}_t}\,
    \hat{\epsilon}_{\theta}(x_t,t,c)}
    {\sqrt{\bar{\alpha}_t}}.
    \label{eq:clean_estimate}
\end{equation}
Thus, the head predicts noise, while the reverse sampling process recovers the continuous measurement.

\subsubsection{Numerical Prediction Post-Processing}
\par
At test time, to enhance prediction accuracy, we further introduce a \emph{coordinate-length consistency} mechanism based on the predicted 5-tuple $\hat{x}_0$, as shown in Fig. \ref{fig:post_proc}. This tuple yields two length estimates: $ \hat{m}_{\text{coord}}$, computed from the predicted coordinates, and $\hat{m}$, directly regressed by the model. We quantify the discrepancy between these two length estimates as
$\Delta=|\hat{m}_{\mathrm{coord}}-\hat{m}|$ and apply
\begin{equation}
    \operatorname{D}_{\tau}(\hat{x}_0)
    =
    \begin{cases}
        \hat{m}, & \Delta>\tau,\\[2pt]
        \dfrac{\hat{m}_{\mathrm{coord}}+\hat{m}}{2},
            & \Delta\leq\tau,
    \end{cases}
    \label{eq:consistency_fusion}
\end{equation}
where the coordinate-length consistency threshold $\tau$ is calibrated on the validation set. For each validation prediction, we compute the absolute discrepancy between the coordinate-derived length and the directly regressed length:
\begin{equation}
\tau = Q_{0.85}\left({\Delta_i}_{i=1}^{N_{\mathrm{val}}}\right),~~\Delta_i =\left|\hat{m}_{\mathrm{coord},i}-\hat{m}_i\right|
\label{eq:threshold}
\end{equation}
where $Q_{0.85}$ denotes the empirical 85th percentile and $N_{\mathrm{val}}$ is the number of validation predictions used for calibration. The percentile level is selected based on validation performance, and the resulting threshold is fixed during testing.

\par
To reduce variability associated with diffusion sampling, we obtain $R=20$ independently initialized predictions~\cite{chan2024estimating}. Each tuple is post-processed separately, and the final scalar length is
\begin{equation}
    (r^\star, m^{\star})
    =
    \operatorname{Median}_{r=1,\ldots,R}
    \left\{
        \operatorname{D}_{\tau}\left(\hat{x}_0^{(r)}\right)
    \right\}.
    \label{eq:ensemble_median}
\end{equation}
where median aggregation applies to post-processed lengths; endpoint predictions are associated with their individual sampled tuples.

\begin{table*}[!t]
    \centering
    \caption{Quantitative comparison on four \texttt{EchoNet-LVH} measurement tasks, including IVSd, LVIDd, LVIDs, and LVPWd (MAE: mean absolute error; RMSE: root mean squared error; $r$: Pearson correlation coefficient; $R^2$: coefficient of determination; MAD: mean absolute deviation reflecting expert measurement variability~\cite{Tromp2022AFV}).}
    \label{tab:comparison_results}
    \setlength{\tabcolsep}{3.5pt}
    \renewcommand{\arraystretch}{1.1}
    \resizebox{\textwidth}{!}{%
    \begin{tabular}{l*{16}{c}}
        \toprule
        \multicolumn{1}{c}{\multirow{2}{*}{\textbf{Models}}}
        & \multicolumn{4}{c}{\textbf{IVSd (MAD: 1.20)}}
        & \multicolumn{4}{c}{\textbf{LVIDd (MAD: 2.97)}}
        & \multicolumn{4}{c}{\textbf{LVIDs (MAD: 3.60)}}
        & \multicolumn{4}{c}{\textbf{LVPWd (MAD: 1.16)}} \\
        \cmidrule(lr){2-5}
        \cmidrule(lr){6-9}
        \cmidrule(lr){10-13}
        \cmidrule(lr){14-17}
        & MAE $\downarrow$ & RMSE $\downarrow$
        & $R^2 \uparrow$ & $r \uparrow$
        & MAE $\downarrow$ & RMSE $\downarrow$
        & $R^2 \uparrow$ & $r \uparrow$
        & MAE $\downarrow$ & RMSE $\downarrow$
        & $R^2 \uparrow$ & $r \uparrow$
        & MAE $\downarrow$ & RMSE $\downarrow$
        & $R^2 \uparrow$ & $r \uparrow$ \\
        \midrule

        VLM (Vicuna)
        & 1.82 & 2.54 & 0.25 & 0.62
        & 3.62 & 5.00 & 0.60 & 0.82
        & 3.88 & 4.99 & 0.65 & 0.82
        & 1.86 & 8.43 & -12.55 & 0.15 \\

        Fine-tuned Qwen3-VL
        & 1.68 & 2.44 & 0.32 & 0.60
        & 3.71 & 4.87 & 0.60 & 0.79
        & 3.82 & 5.01 & 0.64 & 0.81
        & 1.47 & 1.95 & 0.28 & 0.57 \\

        VLM + MLP Regression
        & 1.41 & 2.00 & 0.54 & 0.74
        & 3.00 & 3.95 & 0.74 & 0.86
        & 3.42 & 4.52 & 0.71 & 0.85
        & 1.28 & 1.67 & 0.47 & 0.71 \\

        \midrule
        HAD-VLM (ours)
        & \textbf{1.36} & \textbf{1.96}
        & \textbf{0.56} & \textbf{0.75}
        & \textbf{2.77} & \textbf{3.62}
        & \textbf{0.78} & \textbf{0.89}
        & \textbf{3.05} & \textbf{4.00}
        & \textbf{0.77} & \textbf{0.88}
        & \textbf{1.16} & \textbf{1.54}
        & \textbf{0.55} & \textbf{0.74} \\
        \bottomrule
        \multicolumn{17}{l}{\scriptsize{All MAE, RMSE, and MAD values are reported in mm. Bold values indicate the \textbf{best} reported result for each metric.}}
    \end{tabular}%
    }
\end{table*}

\begin{figure*}[!t]
    \centering 
    \includegraphics[width=0.95\linewidth]{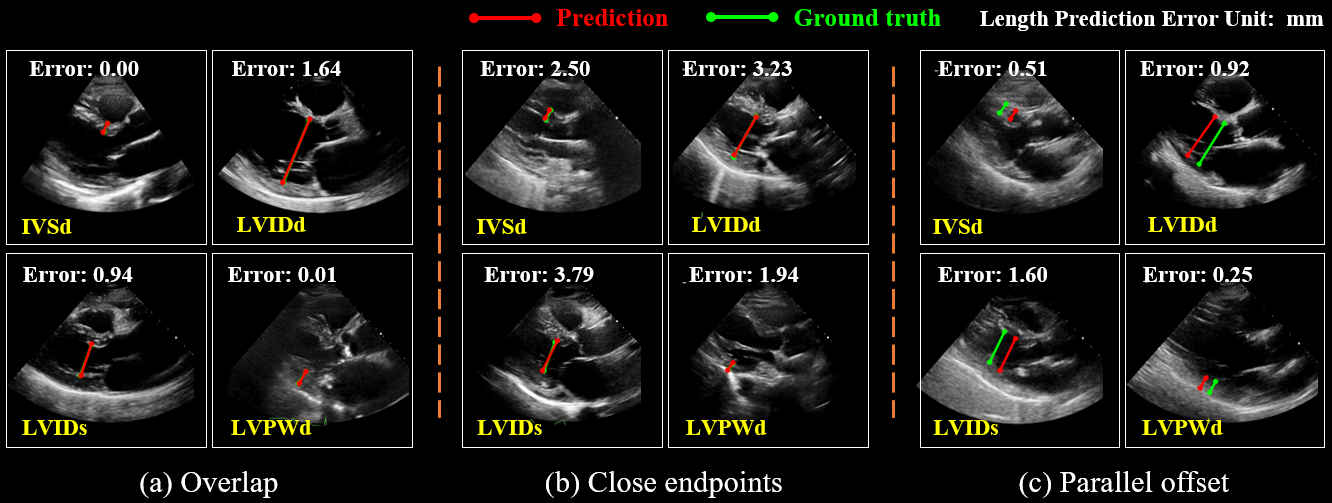}
    \caption{
        Qualitative visualization of three representative prediction patterns.
    }
    \label{fig:visualization_cases}
\end{figure*}

\subsubsection{Hybrid Model Inference}
\par
To decouple numerical regression from autoregressive text decoding without extending the vocabulary, we use square brackets (“\texttt{[}” and “\texttt{]}”) as control tokens, thereby avoiding the need to train embeddings for newly introduced special tokens.
When the model generates “\texttt{[}”, it switches from text generation to diffusion-based regression; when “\texttt{]}” is produced, autoregressive decoding resumes. Specifically, upon generating “\texttt{[}”, we initialize $x_T\sim\mathcal{N}(0, \mathbf{I})$ and perform fixed-step DDIM sampling (4 steps) to obtain the predicted value $\hat{x}_0$. After regression, “\texttt{]}” is manually appended to the sequence, and text decoding continues. This alternating decoding mechanism allows diffusion regression to condition on all preceding visual and textual tokens while avoiding the discretization of continuous numerical values. The final output, therefore, contains both structured textual descriptions and continuous measurement predictions. 

\par
Algorithm~\ref{alg:hadvlm_inference} summarizes the hybrid inference procedure. At each measurement slot, multiple diffusion samples are independently generated and post-processed to obtain an ensemble estimate. The final length and a representative 5-tuple are stored in the measurement records, while the length is also inserted into the text sequence before autoregressive decoding resumes. This procedure preserves structured numerical outputs and makes the predicted measurements available as context for subsequent text generation.

\subsection{Loss Functions and Training Strategy}
\label{subsec:loss_training}
\subsubsection{Loss Functions}
We jointly optimize two objectives: (1) a standard language modeling loss for textual tokens, and (2) a diffusion denoising loss for continuous measurement regression. For text generation, we employ the autoregressive cross-entropy loss: 
\begin{equation}
    \mathcal{L}_{\mathrm{ce}} = - \mathbb{E}_{(v,p,y)} \left[ \sum_{n=1}^{N_y} \log P_{\Theta}\!\left(y_n \mid \mathbf{v},\, p,\,y_{<n}\right) \right],
\end{equation}
where $\mathbf{v}$ and $p$ denote the tokenized visual features and user prompts, respectively, and $y$ represents the target text sequence. For diffusion-based regression, we sample $t$ and noise $\epsilon$ to construct a noisy measurement $x_t=\alpha_t m+\sigma_t\epsilon$. The model predicts the noise component from the LLM hidden state at the diffusion slot: $\hat{\epsilon}_{\theta}(x_t,t,c)=g_{\phi}(h_{\theta}(x_t,t,c))$, where $h_\theta$ denotes the LLM backbone, $g_\phi$ is a lightweight regression head, and $c$ denotes the contextual condition. The diffusion objective minimizes the mean squared error: 
\begin{equation}
    \mathcal{L}_{\mathrm{diff}} = \mathbb{E}_{m,\,t,\,\epsilon}\left[ \left\| \epsilon - \hat{\epsilon}_{\theta}(x_t, t, c) \right\|_2^2 \right].
\end{equation}

\subsubsection{Three-Stage Training Strategy}
We empoly a three-stage training strategy to progressively align visual-language representations, stabilize continuous numerical regression, and ultimately enable joint optimization: 
\begin{enumerate}
    \item \textit{Alignment}: Optimize the visual prompts and the vision-language connector with $\mathcal{L}_{\mathrm{ce}}$ while keeping CLIP and the LLM backbone frozen;
    \item \textit{Regression}: Activate the diffusion segment, optimize diffusion-related modules with $\mathcal{L}_{\mathrm{diff}}$, and apply LoRA to the last four LLM layers;
    \item \textit{Joint Training}: Jointly optimize $\mathcal{L}_{\mathrm{ce}}+\mathcal{L}_{\mathrm{diff}}$ over all trainable parameters to improve consistency between textual descriptions and numerical measurements.
\end{enumerate}

\begin{table}[b]
    \centering
    \caption{\texttt{EchoNet-LVH} dataset split for the measurement tasks.} 
    \label{tab:dataset_split}
    \tabcolsep=8pt
    \resizebox{0.99\columnwidth}{!}{%
    \begin{tabular}{lrrrrr}
        \toprule
        \textbf{Split} & \textbf{IVSd} & \textbf{LVIDd}
        & \textbf{LVIDs} & \textbf{LVPWd} & \textbf{Total} \\
        \midrule
        Training set & 10265 & 10312 & 9859 & 10254 & 40690 \\
        Validation set & 310  & 309  & 284 & 310  & 1213  \\
        Test set & 1163  & 1170  & 1154 & 1164  & 4651  \\
        \bottomrule
    \end{tabular}
    }
\end{table}

\section{Experiments}
\subsection{Experimental Setup}
\subsubsection{Dataset} 
We evaluate the proposed method on \texttt{EchoNet-LVH}~\cite{duffy2022high}, a publicly available ultrasound dataset for quantitative assessment of left ventricular structure. It contains approximately 12,000 B-mode echocardiography videos acquired from the parasternal long-axis view. The dataset provides annotated measurement endpoints in pixel coordinates for three key cardiac structures with phase labels distinguishing diastole and systole. In this work, we focus on four clinically relevant measurements as regression targets: interventricular septum diastole (IVSd), left ventricular internal dimension diastole (LVIDd), left ventricular internal dimension systole (LVIDs), and left ventricular posterior wall diastole (LVPWd). 

\par
As shown in Table~\ref{tab:dataset_split}, the dataset is partitioned into training, validation, and test sets in an 87:3:10 ratio, with 40,690 annotated frames used for training. To facilitate multimodal learning, we convert the structured annotations into text-measurement pairs by wrapping them with diversified natural-language QA prompts generated using ChatGPT, thereby constructing structured textual outputs aligned with the corresponding numerical targets.

\subsubsection{Training Details} 
\par
We jointly train the model to predict IVSd, LVIDd, LVIDs, and LVPWd, with the target measurement specified through the user prompt. Optimization is performed using AdamW with a weight decay of 0.01 on a single NVIDIA A100 GPU, using a batch size of 8 and gradient clipping set to 1.0. Training follows the three-stage schedule described in Sec.~\ref{subsec:loss_training} with training hyperparameters determined empirically based on preliminary experiments: \textit{Stage 1 (Alignment)} is conducted for 15 epochs with a learning rate of $1\times10^{-4}$; \textit{Stage 2 (Regression)} runs for 30 epochs at $1\times10^{-5}$; and \textit{Stage 3 (Joint Training)} proceeds for 15 epochs at $1\times10^{-6}$. During regression and joint training, LoRA is applied to the last four LLM layers with rank $r=32$, scaling factor $\alpha=64$, and dropout rate 0.05. For diffusion-based regression, we adopt $v$-prediction with 100 timesteps and employ the \texttt{squaredcos\_cap\_v2} noise schedule.

\begin{table*}[!t]
\caption{Ablation results across four \texttt{EchoNet-LVH} measurement tasks: IVSd, LVIDd, LVIDs, and LVPWd. (MAE: mean absolute error; RMSE: root mean squared error; $r$: Pearson correlation coefficient; MAD: mean absolute deviation reflecting expert measurement variability~\cite{Tromp2022AFV}. All MAE, RMSE, and MAD values are reported in mm.)}
\label{tab:ablation_results}
\centering
\tabcolsep=0.12cm
\resizebox{0.99\textwidth}{!}{%
\begin{tabular}{lcccccccccccccccc}
\toprule
\multicolumn{1}{c}{\multirow{2}{*}{\textbf{Models}}}
& \multicolumn{4}{c}{\textbf{\scriptsize{IVSd (MAD: 1.20)}}} & \multicolumn{4}{c}{\textbf{\scriptsize{LVIDd (MAD: 2.97)}}} & \multicolumn{4}{c}{\textbf{\scriptsize{LVIDs (MAD: 3.60)}}} & \multicolumn{4}{c}{\textbf{\scriptsize{LVPWd (MAD: 1.16)}}} \\
% \cmidrule(l){2-6} \cmidrule(l){7-11}
\cmidrule(l){2-5} \cmidrule(l){6-9} \cmidrule(l){10-13} \cmidrule(l){14-17}
& {\scriptsize{MAE} $\downarrow$} & {\scriptsize{RMSE} $\downarrow$} & {$R^2\uparrow$} & {$r\uparrow$} % & {\scriptsize{MAD}} 
& {\scriptsize{MAE} $\downarrow$} & {\scriptsize{RMSE} $\downarrow$} & {$R^2\uparrow$} & {$r\uparrow$} & {\scriptsize{MAE} $\downarrow$} & {\scriptsize{RMSE} $\downarrow$} & {$R^2\uparrow$} & {$r\uparrow$} % & {\scriptsize{MAD}} 
& {\scriptsize{MAE} $\downarrow$} & {\scriptsize{RMSE} $\downarrow$} & {$R^2\uparrow$} & {$r\uparrow$} \\ % & {\scriptsize{MAD}} \\
\midrule
w/o prompt tuning & 1.93 & 2.80 & 0.11 & 0.41 % & {} 
                  & 5.07 & 6.64 & 0.26 & 0.58 %\\% & {} \\
                  & 5.60 & 7.40 & 0.22 & 0.54 %& {} 
                  & 1.58 & 2.08 & 0.18 & 0.48 \\%& {} \\
w/o cons. w/o ens. & 1.49 & 2.14 & 0.47 & 0.69 % & {}
                   & 3.04 & 3.98 & 0.75 & 0.87 %\\ %& {} \\
                   & 3.31 & 4.29 & 0.74 & 0.86 %& {}
                   & 1.31 & 1.73 & 0.42 & 0.67 \\%& {} \\
w/ cons. w/o ens. & 1.46 & 2.07 & 0.51 & 0.72 % & {}
                  & 2.96 & 3.85 & 0.75 & 0.87 %\\% & {} \\
                  & 3.34 & 4.41 & 0.72 & 0.85 % & {}
                  & 1.23 & 1.60 & 0.52 & 0.73 \\% & {} \\
% \midrule
% trained w/ single-task & 1.44 & 2.01 & 0.54 & 0.74 % & {}
%             & 3.69 & 4.84 & 0.61 & 0.80 %\\ %& {} \\
%             & 3.69 & 4.89 & 0.66 & 0.83 % & {}
%             & 1.27 & 1.63 & 0.49 & 0.72 \\% & {} \\
\midrule
HAD-VLM (ours) & \textbf{1.36} & \textbf{1.96} & \textbf{0.56} & \textbf{0.75} % & {}
         & \textbf{2.77} & \textbf{3.62} & \textbf{0.78} & \textbf{0.89} %\\ %& {} \\
         & \textbf{3.05} & \textbf{4.00} & \textbf{0.77} & \textbf{0.88} %& {}
         & \textbf{1.16} & \textbf{1.54} & \textbf{0.55} & \textbf{0.74} \\ %& {} \\
\bottomrule
\multicolumn{17}{l}{\scriptsize{w/: with, w/o: without, cons.: coordinate-length consistency, ens.: ensembling strategy. All MAE, RMSE, and MAD values are reported in mm.}}\\
\multicolumn{17}{l}{\scriptsize{Bold values indicate the \textbf{best} reported result for each metric.}}
\end{tabular}
}
\end{table*}

\subsection{Comparison with Baseline Methods}
\par
We compare HAD-VLM with three baselines representing different numerical prediction strategies. \emph{VLM (Vicuna)} uses Vicuna-7B-v1.5 to generate measurement values autoregressively as discrete text tokens. \emph{Fine-tuned Qwen3-VL}~\cite{bai2025qwen3} adopts the same token-based prediction strategy after fine-tuning on \texttt{EchoNet-LVH}, allowing us to examine the effect of using an alternative VLM backbone. \emph{VLM + MLP Regression} replaces the diffusion-based prediction head with an MLP that directly regresses continuous measurement values, which shares the same settings of REO-VLM~\cite{xue2024reo}. All methods are evaluated on the same test samples for each measurement task.

\par
As shown in Table~\ref{tab:comparison_results}, HAD-VLM consistently outperforms the Vicuna-based autoregressive baseline across all four measurements. For IVSd, MAE decreases from 1.82 to 1.36~mm, corresponding to a 25.3\% reduction, while $R^2$ increases from 0.25 to 0.56. For LVIDd, MAE decreases from 3.62 to 2.77~mm, a 23.5\% reduction, and $R^2$ improves from 0.60 to 0.78. Similar improvements are observed for LVIDs and LVPWd, with MAE decreasing from 3.88 to 3.05~mm and from 1.86 to 1.16~mm, respectively. These results support the effectiveness of the proposed hybrid framework for quantitative ultrasound measurement.

\par
To further examine the roles of the backbone and numerical prediction strategy, we compare HAD-VLM with fine-tuned Qwen3-VL and a VLM equipped with a direct MLP regression head. Qwen3-VL generates numerical values autoregressively as discrete tokens. HAD-VLM achieves lower MAE across all four tasks, including 2.77 and 3.05~mm for LVIDd and LVIDs, compared with 3.71 and 3.82~mm for Qwen3-VL. Thus, adopting an alternative VLM backbone while retaining token-based numerical prediction does not match the accuracy of HAD-VLM under the evaluated settings. Direct MLP regression improves upon both autoregressive baselines, but HAD-VLM further reduces LVIDd and LVIDs MAE from 3.00 to 2.77~mm and from 3.42 to 3.05~mm, corresponding to reductions of 7.7\% and 10.8\%, respectively. These improvements, together with lower RMSE and higher $R^2$ and Pearson correlation across all four tasks, favor the diffusion-based framework over direct regression in the evaluated configurations.

\par
The MAEs achieved by HAD-VLM are of a similar magnitude to previously reported expert measurement variability: 1.20~mm for IVSd, 2.97~mm for LVIDd, 3.60~mm for LVIDs, and 1.16~mm for LVPWd reported by Tromp \emph{et al.}~\cite{Tromp2022AFV}. This comparison provides context for the observed errors, although differences in datasets and evaluation protocols prevent direct conclusions about equivalence to expert performance or the contribution of annotation variability to the remaining errors.

\par
Fig.~\ref{fig:visualization_cases} illustrates the relationship between spatial localization and numerical accuracy. In panel~(a), predicted segments closely overlap with the reference annotations and yield small length errors. Panel~(b) shows cases in which endpoint deviations alter the estimated segment length, resulting in larger measurement errors. In panel~(c), spatially displaced segments nevertheless have lengths close to those of the reference segments and therefore yield small numerical errors. These examples highlight that accurate length prediction does not necessarily imply anatomically accurate caliper placement, motivating assessment of both endpoint localization and numerical measurement.

\subsection{Ablation Studies}
\label{subsec:ablation}
\par
We evaluate the contributions of visual prompt tuning and the two components of the proposed post-processing strategy: \textit{coordinate-length consistency} and ensemble prediction. Table~\ref{tab:ablation_results} compares the complete HAD-VLM with three variants: removing visual prompt tuning (\emph{w/o prompt tuning}), disabling both post-processing components (\emph{w/o cons. w/o ens.}), and retaining \textit{coordinate-length consistency} while disabling ensemble prediction (\emph{w/ cons. w/o ens.}).

\par
Removing visual prompt tuning (\emph{w/o prompt tuning}) leads to the largest performance degradation among the evaluated variants, with deterioration across all four tasks and metrics. For IVSd, MAE increases from 1.36 to 1.93~mm, while $R^2$ decreases from 0.56 to 0.11. The degradation is particularly pronounced for ventricular diameter measurements: LVIDd and LVIDs MAE increase from 2.77 to 5.07~mm and from 3.05 to 5.60~mm, respectively. These results highlight the importance of adapting the pretrained CLIP encoder to ultrasound images for accurate anatomical measurement.

\par
Disabling both post-processing components (\emph{w/o cons. w/o ens.}) also reduces performance. For example, LVIDd MAE increases from 2.77 to 3.04~mm (9.7\%), accompanied by a decrease in $R^2$ from 0.78 to 0.75. Introducing coordinate-length consistency without ensembling (\emph{w/ cons. w/o ens.}) reduces LVPWd MAE from 1.31 to 1.23~mm (6.1\%). Similar improvements are observed for IVSd and LVIDd, although LVIDs MAE increases slightly from 3.31 to 3.34~mm. This suggests that consistency-based refinement benefits most measurement targets, with its effectiveness varying across tasks.

\par
Adding ensemble prediction to the consistency-enabled variant improves all four metrics across all tasks. For example, LVIDd MAE decreases from 2.96 to 2.77~mm, while LVIDs MAE decreases from 3.34 to 3.05~mm. Their corresponding $R^2$ values increase from 0.75 to 0.78 and from 0.72 to 0.77, respectively. These results support aggregating repeated diffusion predictions to improve measurement accuracy.

\par
Finally, the improvement over autoregressive prediction is already evident without either post-processing component. For LVIDd, the hybrid model without consistency or ensembling achieves an MAE of 3.04~mm, compared with 3.62~mm for the autoregressive baseline in Table~\ref{tab:comparison_results}; the complete framework further reduces this error to 2.77~mm. This comparison indicates that post-processing provides additional gains beyond those achieved by the hybrid prediction framework.

\begin{table}[!t]
    \centering
    \caption{Comparison of single-task and multi-task training on \texttt{EchoNet-LVH}.}
    \label{tab:multitask_results}
    \setlength{\tabcolsep}{6pt}
    \renewcommand{\arraystretch}{1.1}
    \resizebox{\columnwidth}{!}{%
    \begin{tabular}{llcccc}
        \toprule
        \textbf{Task} & \textbf{Training}
        & \textbf{MAE $\downarrow$}
        & \textbf{RMSE $\downarrow$}
        & \textbf{$R^2 \uparrow$}
        & \textbf{$r \uparrow$} \\
        \midrule

        \multirow{2}{*}{IVSd}
        & Single-task & 1.44 & 2.01 & 0.54 & 0.74 \\
        & HAD-VLM (ours)
        & \textbf{1.36} & \textbf{1.96}
        & \textbf{0.56} & \textbf{0.75} \\
        \midrule

        \multirow{2}{*}{LVIDd}
        & Single-task & 3.69 & 4.84 & 0.61 & 0.80 \\
        & HAD-VLM (ours)
        & \textbf{2.77} & \textbf{3.62}
        & \textbf{0.78} & \textbf{0.89} \\
        \midrule

        \multirow{2}{*}{LVIDs}
        & Single-task & 3.69 & 4.89 & 0.66 & 0.83 \\
        & HAD-VLM (ours)
        & \textbf{3.05} & \textbf{4.00}
        & \textbf{0.77} & \textbf{0.88} \\
        \midrule

        \multirow{2}{*}{LVPWd}
        & Single-task & 1.27 & 1.63 & 0.49 & 0.72 \\
        & HAD-VLM (ours)
        & \textbf{1.16} & \textbf{1.54}
        & \textbf{0.55} & \textbf{0.74} \\
        \bottomrule
        \multicolumn{6}{l}{\scriptsize{MAE and RMSE are reported in mm. Bold values indicate the \textbf{best} result.}}
    \end{tabular}%
    }
\end{table}

\subsection{Single-Task versus Multi-Task Training}
\label{subsec:multitask}
\par
To examine the effect of joint training across measurement tasks, we compare a single multi-task HAD-VLM with four independently trained single-task models, each dedicated to one measurement. In the multi-task setting, the target measurement is specified through the user prompt, allowing all four tasks to be supported by a shared model. As shown in Table~\ref{tab:multitask_results}, multi-task training achieves lower MAE and RMSE and higher $R^2$ and Pearson correlation across all four measurements. The largest improvements are observed for ventricular diameter estimation. For LVIDd, MAE decreases from 3.69 to 2.77~mm, corresponding to a 24.9\% reduction, while RMSE decreases from 4.84 to 3.62~mm and $R^2$ increases from 0.61 to 0.78. For LVIDs, MAE decreases from 3.69 to 3.05~mm, a 17.3\% reduction, accompanied by an RMSE decrease from 4.89 to 4.00~mm and an increase in $R^2$ from 0.66 to 0.77. Pearson correlation also improves from 0.80 to 0.89 for LVIDd and from 0.83 to 0.88 for LVIDs.

\par
Smaller but consistent gains are observed for wall thickness measurements. Multi-task training reduces IVSd and LVPWd MAE from 1.44 to 1.36~mm and from 1.27 to 1.16~mm, respectively, with corresponding improvements in the remaining metrics. These findings are consistent with beneficial representation sharing across related anatomical measurement tasks, with no observed performance degradation on any evaluated task. Joint training therefore enables HAD-VLM to support multiple prompt-specified measurements within a single model while improving accuracy.

\subsection{Evaluation on Language Ability Preservation}
\label{subsec:medical_qa}
\par
To assess whether adaptation to quantitative ultrasound measurement affects language performance, we evaluate HAD-VLM and the original Vicuna-7B-v1.5 checkpoint on the medical question answering dataset from~\cite{11123690}. We report word-level F1 and BERTScore, including precision, recall, and F1, to assess lexical overlap and semantic similarity between generated and reference answers.

\par
As shown in Table~\ref{tab:qa_results}, HAD-VLM achieves higher scores across all four metrics. Word-level F1 increases from 0.298 to 0.320, while BERTScores (Precision/Recall/F1) improve from 0.855, 0.885, and 0.869 to 0.873, 0.891, and 0.882, respectively. These results indicate that training on the curated \texttt{EchoNet-LVH} dataset preserves medical question answering performance on the evaluated benchmark while introducing continuous measurement capabilities.

\subsection{Cross-Modality Generalization to X-Ray Imaging}
% \subsection{Evaluation on X-Ray Images}
\label{subsec:xray}
\subsubsection{Dataset}
\par
\begin{figure}[!t]
    \centering 
    \includegraphics[width=0.85\linewidth]{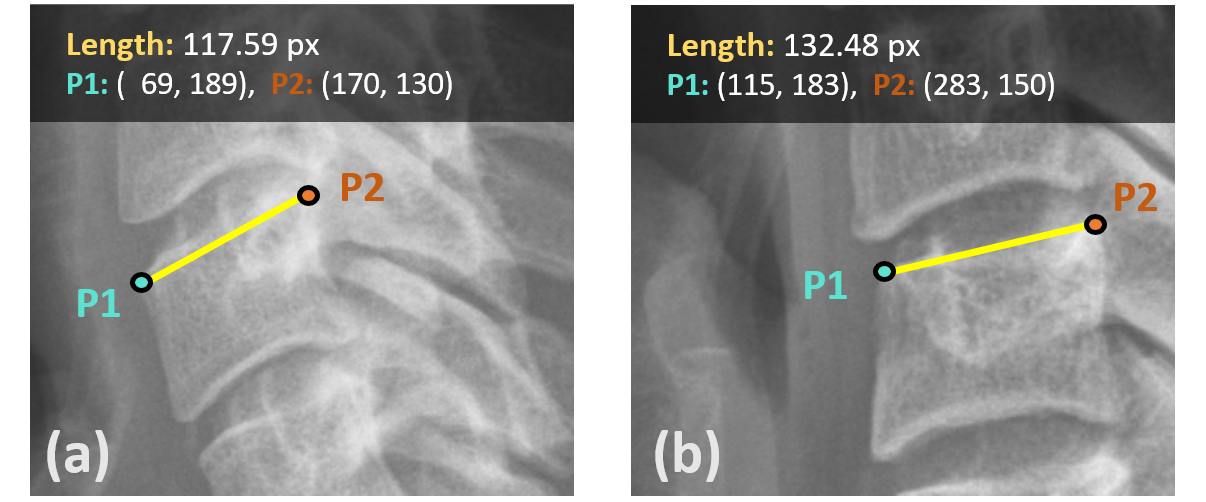}
    \caption{Illustration of superior endplate length measurements for the \textit{Left:} (a) C4 and \textit{Right:} (b) C5 vertebrae.}
    \label{fig:xray_samples}
\end{figure}

To assess the applicability of HAD-VLM beyond ultrasound, we evaluate superior endplate length measurement in cervical spine radiographs. We construct a dataset of 9,926 samples from the Cervical Spine X-ray Atlas (\texttt{CSXA}) V3.0~\cite{ran2024high} by cropping C4 and C5 vertebral regions, with an 85:15 training--test split. As illustrated in Fig.~\ref{fig:xray_samples}, the target measurement is defined as the Euclidean distance between the anterior and posterior endpoints of the superior endplate of the vertebral body. Since the images originate from heterogeneous sources and consistent pixel-to-physical-distance calibration is unavailable, measurements are evaluated in pixel space without applying physical-distance calibration; accordingly, all lengths and errors are reported in pixels.

\begin{table}[t]
% \begin{table}[!t]
\centering
\caption{Evaluation results on the medical QA dataset~\cite{11123690}.}
\label{tab:qa_results}
\begingroup
\small
\setlength{\tabcolsep}{6pt}
\renewcommand{\arraystretch}{1.0}
\resizebox{\columnwidth}{!}{%
    \begin{tabular}{lcccc}
        \toprule
        \textbf{Models} & \textbf{F1 $\uparrow$} & \textbf{BERT P $\uparrow$} & \textbf{BERT R $\uparrow$} & \textbf{BERT F1 $\uparrow$} \\
        \midrule
        VLM (vicuna)   & 0.298 & 0.855 & 0.885 & 0.869 \\
        HAD-VLM (ours) & \textbf{0.320} & \textbf{0.873} & \textbf{0.891} & \textbf{0.882} \\
        % with co-training   & \textbf{0.339} & 0.850 & 0.857 & 0.853 \\
        \bottomrule
        \multicolumn{5}{l}{\scriptsize{Bold values indicate the \textbf{best} reported result for each metric.}}
    \end{tabular}
}
\endgroup
\end{table}

% \begin{table}[h]
%     \centering
%     \caption{\texttt{CSXA} dataset split for the measurement task.} 
%     \label{tab:xray_split}
%     \tabcolsep=14pt
%     \resizebox{0.99\columnwidth}{!}{%
%     \begin{tabular}{lrrr}
%         \toprule
%         \textbf{Split} & \textbf{C4} Region & \textbf{C5} Region & \textbf{Total} \\
%         \midrule
%         Training set & 745 & 745 & 1490 \\
%         Test set & 4218  & 4218 & 8436  \\
%         \bottomrule
%     \end{tabular}
%     }
% \end{table}

\subsubsection{Results}
\par
The reference superior endplate lengths in the curated dataset have a mean of 170.4~pixels and a standard deviation of 29.73~pixels. As shown in Table~\ref{tab:xray_results}, HAD-VLM reduces the mean absolute error from 11.4 to 6.7~pixels compared with the Vicuna-based autoregressive baseline, corresponding to a 41.2\% reduction. The standard deviation of the absolute errors also decreases from 93.9 to 5.7~pixels, indicating substantially less dispersion in measurement errors. These results demonstrate improved accuracy and consistency for C4/C5 superior endplate length estimation and support the applicability of the proposed framework beyond ultrasound.

\begin{table}[h]
    \centering
    \caption{Measurement results on the \texttt{CSXA}-derived X-ray dataset.}
    \label{tab:xray_results}
    \setlength{\tabcolsep}{10pt}
    \renewcommand{\arraystretch}{1.1}
    \begin{tabular}{llc}
        \toprule
        \textbf{Models} & \textbf{Numerical prediction} & \textbf{Absolute errors} \\
        \midrule
        VLM (vicuna) & Discrete tokens & 11.4$\pm$93.9 \\
        HAD-VLM (ours) & Continuous diffusion & \textbf{6.7}$\pm$\textbf{5.7} \\
        \bottomrule
        \multicolumn{3}{l}{\scriptsize{Absolute errors are reported in mean$\pm$std pixels.}} \\
        \multicolumn{3}{l}{\scriptsize{Bold values indicate the \textbf{best} reported result.}}
    \end{tabular}
\end{table}

\section{Discussion}
\par
In this work, we investigate quantitative ultrasound measurement and move beyond task-specific regression networks by introducing a hybrid vision-language framework, HAD-VLM, that generates text autoregressively and predicts numerical measurements directly in continuous space via an embedded diffusion-based process. Experimental results demonstrate improved regression accuracy and robustness compared with the discrete token-based autoregressive baseline. A current limitation is that the framework focuses on linear measurements. Nevertheless, such measurements are widely applicable in clinical practice, including tumor size estimation, vessel diameter assessment, organ length measurement (e.g., kidney length), myocardial wall thickness evaluation, etc. Extending the proposed framework to more complex geometric or area-based measurements is a promising direction for future work.

\par
Beyond ultrasound, the proposed hybrid autoregressive-diffusion formulation can potentially apply to other imaging modalities, such as X-ray, CT, and MRI. Another promising direction is to incorporate higher-quality report supervision that follows a “qualitative-to-quantitative” workflow, where anatomical structures and boundary characteristics are described before numerical measurements are reported. Such structured supervision may provide meaningful anatomical priors, thereby improving the accuracy and reliability of quantitative prediction.

\section{Conclusion}
\par
We presented HAD-VLM, a unified vision-language framework for prompt-conditioned quantitative ultrasound measurement that combines autoregressive text generation with diffusion-based continuous prediction. A five-tuple representation couples caliper endpoint coordinates with measurement length, while a post-processing strategy comprising coordinate-length consistency and ensemble prediction further refines the estimates. On four EchoNet-LVH measurement tasks, HAD-VLM achieved MAEs of 1.36, 2.77, 3.05, and 1.16~mm for IVSd, LVIDd, LVIDs, and LVPWd, respectively, outperforming both autoregressive baselines and direct MLP regression across all evaluated metrics. Ablation experiments supported the contributions of ultrasound-specific visual adaptation and post-processing, while multi-task training improved performance over independently trained single-task models. Medical question answering performance was preserved on the evaluated benchmark. An additional evaluation of cervical vertebral superior endplate length measurement reduced MAE from 11.4 to 6.7~pixels, providing preliminary evidence of applicability beyond ultrasound. These findings support integrating language generation and continuous numerical prediction within a shared architecture to enable flexible, anatomically grounded measurement. HAD-VLM offers a foundation for language-guided measurement assistance and the incorporation of quantitative image findings into clinical reporting workflows. Future work will investigate validation across institutions and imaging conditions, measurement uncertainty estimation, and integration into comprehensive report generation.

\bibliographystyle{IEEEtran}
\bibliography{IEEEabrv, reference}

\end{document}